\documentclass{article}

\usepackage{microtype}
\usepackage{graphicx}
\usepackage{subcaption}
\usepackage{booktabs} 
\usepackage{booktabs}
\usepackage{multirow}
\usepackage{array}
\usepackage{xcolor}
\usepackage{makecell}
\usepackage{hyperref}

\usepackage[accepted]{icml2026}

\usepackage{amsmath}
\usepackage{amssymb}
\usepackage{mathtools}
\usepackage{amsthm}
\usepackage{array}
\usepackage{booktabs}
\usepackage{multirow}
\usepackage{makecell}
\usepackage{array}
\usepackage{multirow}
\usepackage{makecell}
\usepackage{booktabs}

\newcolumntype{M}[1]{>{\centering\arraybackslash}m{#1}}
\newcolumntype{C}[1]{>{\centering\arraybackslash}m{#1}}

\usepackage[capitalize,noabbrev]{cleveref}

\theoremstyle{plain}

\theoremstyle{definition}

\theoremstyle{remark}

\usepackage[disable,textsize=tiny]{todonotes}
\icmltitlerunning{Efficient Diffusion LLMs via TSPD and Confidence Extrapolation}

\begin{document}

\twocolumn[
  \icmltitle{Efficient Diffusion LLMs via Temporal-Spatial Parallel Decoding and Confidence Extrapolation}

  \icmlsetsymbol{equal}{*}

  \begin{icmlauthorlist}
  \icmlauthor{Zekai Li}{yyy}
    \icmlauthor{Ji Liu}{yyy}
    \icmlauthor{Yiqing Huang}{yyy}
    \icmlauthor{Ziqiong Liu}{yyy}
    \icmlauthor{Dong Li}{yyy}
    \icmlauthor{Emad Barsoum}{yyy}
  \end{icmlauthorlist}

  \icmlaffiliation{yyy}{Advanced Micro Devices, Inc. (AMD)}

  \icmlcorrespondingauthor{Ji Liu}{liuji@amd.com}
    \icmlcorrespondingauthor{Ziqiong Liu}{ziqioliu@amd.com}

  \icmlkeywords{Machine Learning, ICML}

  \vskip 0.3in
]



\printAffiliationsAndNotice{}  

\begin{abstract}
Diffusion-based large language models (dLLMs) support parallel text generation via iterative denoising, yet inference remains latency-heavy because many steps are spent on redundant refinement and repeated remasking of tokens whose final values are already determined.
Prior acceleration methods mainly depend on step-local confidence heuristics or fixed schedules, which are sensitive to prompt and task variation and ignore strong positional effects within a sequence.
We cast diffusion decoding as a dynamic control problem and show that token-wise denoising trajectories provide the key signal for reliable control.
We propose a trace-aware decoding framework with two components.
First, \textbf{Temporal-Spatial Parallel Decoding (TSPD)} uses a lightweight temporal-spatial controller that consumes per-token trajectory features, including confidence, entropy, and momentum, together with token position, to decide when a token has converged and can be safely fixed.
Second, we introduce \textbf{Confidence Extrapolation (CE)}, a training-free state-space module that forecasts future logit trends with uncertainty to support proactive decisions, including safe look-ahead and targeted stabilization when trajectories are oscillatory or underconfident.
Together, TSPD and CE reduce unnecessary denoising iterations while preserving output quality, and they compose cleanly with system optimizations such as KV caching.

\end{abstract}

\section{Introduction}
\label{sec:intro}

Diffusion-based large language models (dLLMs)~\cite{nie2025large,ye2025dream7b,khanna2025mercury,zhao2025d1} offer an alternative to autoregressive decoding~\cite{schuurmans2024autoregressive,radford2018improving} by updating many token positions in parallel through iterative denoising.
This framework supports flexible generation orders, infilling, and a direct quality speed trade off through the number of denoising steps~\cite{ho2020denoising,song2020denoising,nichol2021improved}.
Recent masked diffusion Language models~\cite{sahoo2024simple,he2023diffusionbert,nie2024scaling} and LLaDA style models~\cite{nie2025large} show that diffusion can be competitive at scale.
Despite this parallelism, inference latency remains high in practice.
Each denoising step typically requires a full sequence forward pass under bidirectional attention~\cite{seo2016bidirectional,wibisono2023bidirectional,feng2025learning}, and many steps are spent refining tokens whose final values are already determined.
Repeated remasking further amplifies this cost when uncertainty persists at a small subset of positions.

Prior work reduces this overhead by reusing computation across steps, adjusting unmasking schedules, or committing more tokens per step~\cite{wu2025fastdllmtrainingfreeaccelerationdiffusion,wu2025fastv2,kong2025accelerating,wang2025diffusion,israel2025accelerating}.
These methods improve throughput, but most commit and stopping decisions still depend on step local heuristics, fixed schedules, or global thresholds~\cite{li2025diffusion,chen2025dparallel,chen2025beyond}.
Such signals are brittle because diffusion confidence is not uniformly calibrated across prompts, domains, and token positions.
Empirically, diffusion decoding exhibits structured temporal and spatial behaviors, including delayed stabilization, underconfident but correct tokens, and oscillatory confidence~\cite{huang2025pc,wang2025time}.
A single step snapshot or heuristic rule cannot separate these regimes.
As a result, local rules often pay for redundant denoising after convergence, or they commit prematurely and lose quality.

In this work, we treat diffusion decoding as a \emph{sequential control problem} rather than a sequence of independent threshold tests.
A key observation is that denoising produces token wise \emph{trajectories} across steps, such as confidence and entropy traces, whose temporal patterns reveal both correctness and stability.
This shifts the question from ``is the token confident at step $t$'' to ``does the trajectory indicate convergence, and how will it change if we continue denoising.''
Answering this requires a temporal-spatial-aware controller that can interpret patterns such as steady growth, oscillation, and delayed stabilization, as well as a mechanism that can exploit predictable trends to avoid unnecessary steps.

We propose \textbf{Temporal-Spatial Parallel Decoding (TSPD)}, a temporal-spatial-aware framework that improves both efficiency and accuracy.
TSPD introduces a lightweight \emph{temporal-spatial controller} that predicts whether a position is ready to be fixed using sequential trace features (confidence, entropy, and momentum over recent steps) together with relative spatial position within a block.
A shared sequential perception model captures dynamics that step local classifiers and hand tuned thresholds often miss, improving robustness across prompts and positional effects.
To reduce waiting time when confidence is rising but not yet decisive, we further introduce a training-free and plug-in \textbf{Confidence Extrapolation mechanism (CE)}.
CE models confidence evolution as a simple state space process and forecasts near future confidence together with uncertainty.
A risk-aware horizon rule limits look ahead based on left context progress and forecast uncertainty, so extrapolation is applied only when supported by reliable history.

We evaluate our framework (TSPD and CE) on LLaDA-8B-Instruct~\cite{nie2025large} across four benchmarks spanning arithmetic reasoning and code generation, including GSM8K~\cite{cobbe2021training}, MATH~\cite{hendrycks2021measuring}, HumanEval~\cite{chen2021evaluating}, and MBPP~\cite{austin2021program}.
Across multiple dLLM backbones and generation settings, our method delivers a consistently stronger speed quality tradeoff than state of the art baselines.
On GSM8K with 256 token generation, it achieves $5.0\times$ speedup without cache, and compounds to $11.2\times$ when combined with KV caching, showing that our controller level gains are complementary to system optimizations.
The advantage increases with longer generation: at 1024 tokens, our approach attains 64.1 Tokens Per Second (TPS) and up to $58.3\times$ speedup, while keeping accuracy essentially unchanged and in some cases slightly improved.
In summary, our contributions are threefold:

\begin{itemize}
    \item We propose \textbf{TSPD}, a temporal spatial aware decoding controller that uses temporal confidence dynamics and positional context to make robust token fixing decisions, mitigating the brittleness of step local heuristics.
    \item We introduce a training free and plug and play \textbf{confidence extrapolation} module that forecasts near future confidence under a state space model with quantified uncertainty, and applies risk aware look ahead to complement diverse decoding controllers.
    \item We provide extensive evaluations and ablations on a wide range of dLLMs, task domains, and generation settings, demonstrating substantial speedups with near lossless accuracy and full compatibility with KV cache.
\end{itemize}

\section{Related Work}

\subsection{Diffusion-Based Large Language Models}
Diffusion models~\cite{sohl2015deep,song2019generative} have been extended from continuous data to discrete sequences, enabling non-autoregressive generation via iterative denoising.
Early discrete diffusion work established Markov formulations in categorical spaces, including multinomial diffusion~\cite{hoogeboom2021argmaxflows} and D3PM~\cite{austin2021structured}.
Continuous-time views~\cite{campbell2022continuous} further unify discrete diffusion and support flexible sampling schedules.
SEDD~\cite{lou2023discrete} improves learning by modeling ratios of marginal probabilities with a score-entropy objective, narrowing the gap to autoregressive language models (LMs).

Building on these foundations, masked dLLMs have become a practical paradigm for text generation.
MDLM~\cite{sahoo2024simple} shows that a simplified masked-diffusion recipe can substantially improve perplexity and sampling quality.
At scale, diffusion LLMs are developed either by adapting strong autoregressive backbones (e.g., DiffuLLaMA)~\cite{gong2024scalingdiffusionlanguagemodels} or by training from scratch (e.g., LLaDA)~\cite{nie2025large}, achieving competitive instruction following and reasoning.
Dream 7B~\cite{ye2025dream7b} further advances open diffusion LLMs and highlights diffusion-specific capabilities such as arbitrary-order generation and infilling with tunable quality--speed trade-offs.

\begin{figure}[t]
    \centering
    \begin{minipage}[h]{0.48\linewidth}
        \centering
      \includegraphics[width=\linewidth]{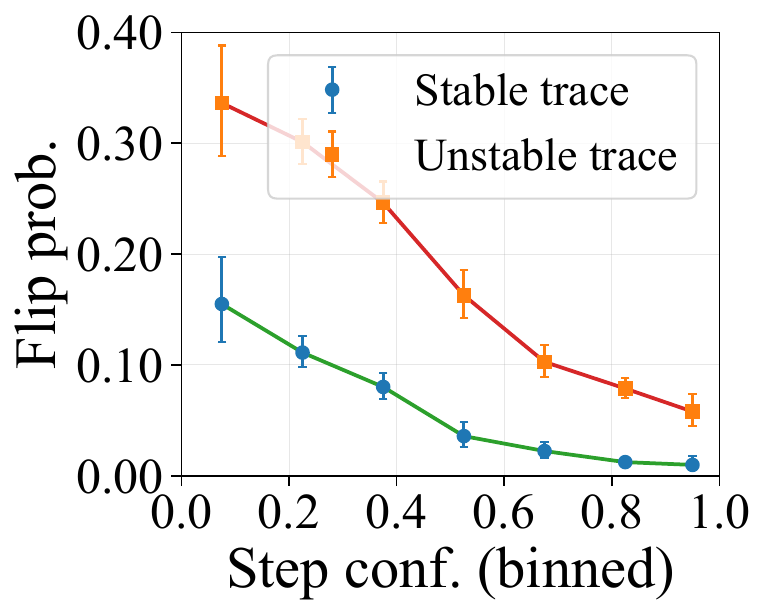}
        \caption{Same local confidence with different trace can lead to different decoding results.}
        \label{fig:temp}
        \vspace{-5mm}
    \end{minipage}%
    \hfill
    \begin{minipage}[h]{0.48\linewidth}
        \centering
        \includegraphics[width=\linewidth]{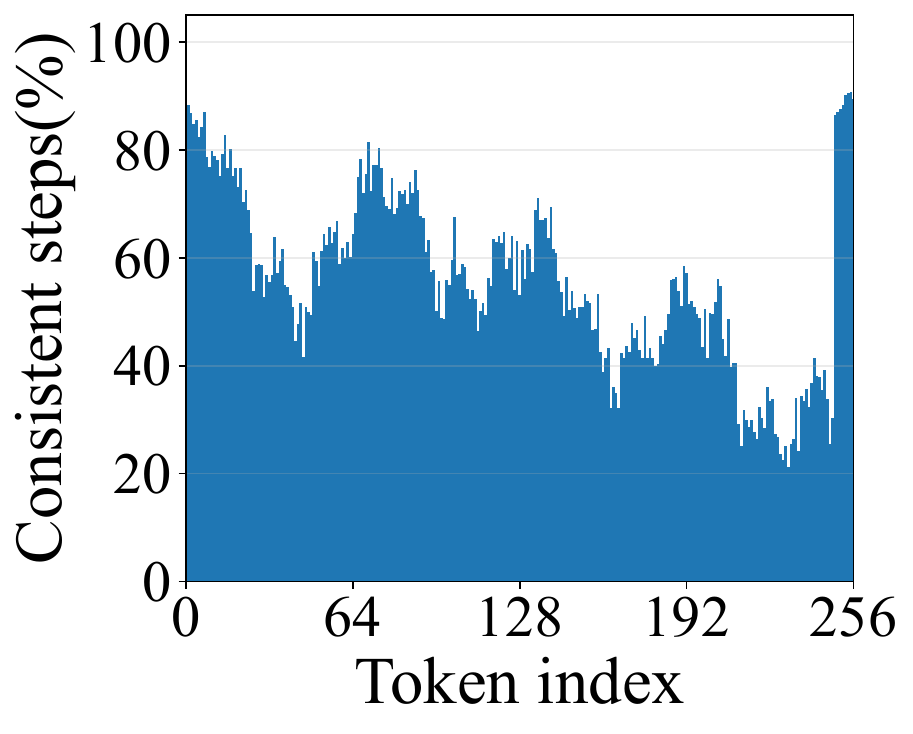}
        \caption{Tokens at more rightward spatial positions tend to stabilize later.}
        \label{fig:spatial}
        \vspace{-5mm}
    \end{minipage}
        \vspace{-0mm}
\end{figure}

\subsection{Decoding Acceleration for Diffusion LLMs}
Despite parallel updates, diffusion decoding can be latency-heavy because each denoising step typically requires a full-sequence forward pass under bidirectional attention.
Prior work improves efficiency via three main directions: (i) computation reuse and caching, (ii) better parallel commitment or sampling schedules, and (iii) plug-in accelerators.

On computation reuse, dLLM-Cache~\cite{liu2025dllmcache} caches static prompts and selectively updates stable tokens; Fast-dLLM~\cite{wu2025fastdllmtrainingfreeaccelerationdiffusion} introduces block-wise approximate KV caching with confidence-aware parallel decoding; DInfer~\cite{ma2025dinfer} modularizes the inference pipeline and reports $\sim$10$\times$ speedup over Fast-dLLM without quality loss; Fast-dLLM v2~\cite{wu2025fastv2} further supports block-diffusion adaptation of pretrained autoregressive backbones and hierarchical KV caching, achieving up to $2.5\times$ speedup over standard autoregressive decoding.
On algorithmic decoding, SlowFast Sampling~\cite{wei2025slowfast} alternates conservative exploration and accelerated decoding; Local Leap~\cite{kong2025accelerating} exploits local determinism around high-confidence anchors; Learn2PD~\cite{bao2025learn2pd} learns a parallel decoding policy with strong speedups on LLaDA; and adaptive parallel decoding~\cite{israel2025accelerating} dynamically controls parallelism and uses a small autoregressive verifier for a controllable speed--quality trade-off.
Training-free plug-ins include FlashDLM~\cite{hu2025flashdlm} (FreeCache + guided diffusion), ES-dLLM~\cite{anonymous2026esdllm} (early skipping via convergence signals), and foreseeing movement decoding~\cite{mo2025decoding}.

Most existing methods still make decisions from per-step heuristics or fixed schedules and respond to uncertainty by passively running more denoising steps.
In contrast, we emphasize that diffusion decoding exposes informative \emph{token-wise trajectories} (confidence/probability traces) that enable more input-adaptive control.
Accordingly, we develop (i) a temporal--spatial correctness sensor that fuses history features with positional cues to infer stabilization, and (ii) an \emph{active forecasting} module that extrapolates near-future confidences with calibrated uncertainty (e.g., via state-space/Kalman-style models~\cite{kalman1960new}) for risk-limited look-ahead decisions.
This trajectory-aware perspective complements caching and sampler optimizations and targets finer-grained speed--quality control for diffusion LLM inference.

\begin{figure}[t]
    \centering
    \begin{minipage}[t]{0.48\linewidth}
        \centering
        \includegraphics[width=\linewidth]{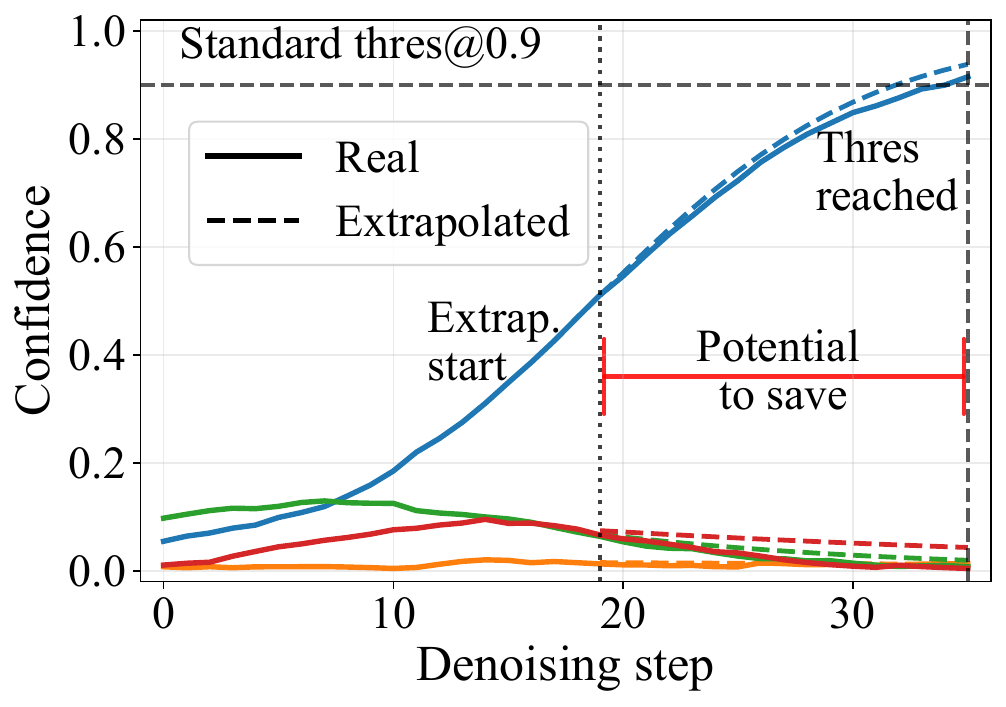}
        \caption{Missed acceleration opportunity of passive waiting compared to lookahead.}
        \label{fig:extrap1}
        \vspace{-5mm}
    \end{minipage}%
    \hfill
    \begin{minipage}[t]{0.48\linewidth}
        \centering
        \includegraphics[width=\linewidth]{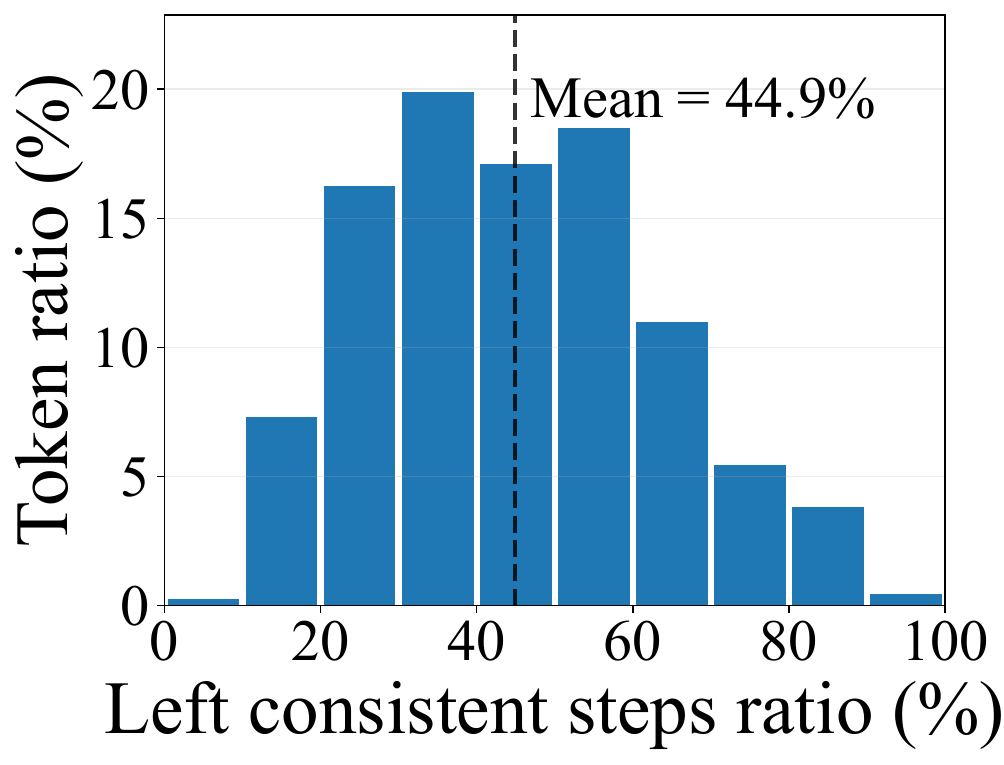}
        \caption{Distribution of consistent steps ratio that can be extrapolated.}
        \label{fig:extrap2}
        \vspace{-5mm}
    \end{minipage}
\end{figure}

\begin{figure*}[!t]
    \centering
    \includegraphics[width=\linewidth]{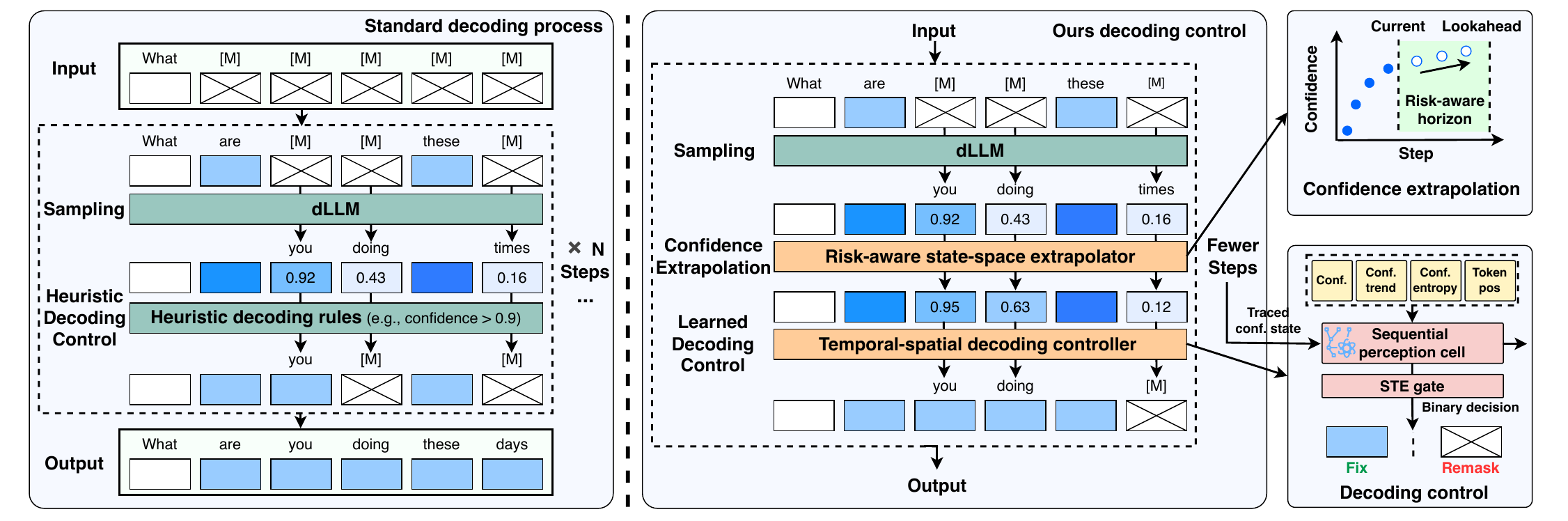}
    \caption{Comparison between standard dLLM parallel decoding (left) and our framework (right).
Standard decoding applies step-local heuristics after each full-sequence denoising pass, often revisiting already-correct tokens.
Our framework inserts a risk-aware confidence extrapolator (CE) and a temporal-spatial decoding controller (TSPD): CE provides uncertainty-aware look-ahead confidence, and TSPD uses token-wise confidence traces and position context to fix stable tokens.
This reduces redundant steps while preserving quality.}

    \vspace{-2mm}
    \label{caption}
\end{figure*}

\section{Methods}

\subsection{Preliminary}
\label{sec:preliminary}

\subsubsection{Parallel Decoding of Diffusion LLMs}
\label{sec:diffusion_llm}

Diffusion LLMs (dLLMs) generate a length-$L$ sequence $x_0 \in \{0,1,\ldots,V-1\}^{L}$ by iteratively denoising a masked version of the sequence.
Let $\alpha \in [0,1]$ denote the masking ratio (noise level) to avoid overloading $t$ as ``time.''
The forward process independently replaces each token with the mask token $m$ with probability $\alpha$:
\begin{equation}
q_{\alpha}(x^\alpha\!\mid x)=\prod_{i=1}^{L}\big[(1-\alpha)\mathbf{1}[x_i^\alpha=x_i]+\alpha\,\mathbf{1}[x_i^\alpha=m]\big].
\label{eq:forward}
\end{equation}
where $x_\alpha$ is the corrupted sequence at masking ratio $\alpha$.
The reverse process denoises in $K$ discrete steps by repeatedly predicting a token distribution for masked positions.
Given a prompt $c=(c_1,\ldots,c_M)$ and current masked response $y^{(k)}$, a mask predictor $p_\theta$ outputs per-position distributions and we form a provisional sequence by greedy decoding:
\begin{equation}
\hat y^{(k)}_i=\arg\max_{v\in[V]}p_\theta\!\left(v\,\middle|\,c,y^{(k)}\right),\quad i\in[L].
\label{eq:greedy}
\end{equation}
and remask the $n_k$ positions with the smallest $c_i$, where $n_k$ (equivalently the masking ratio) follows a predefined schedule that decreases over steps.
This parallel denoising--remasking loop yields the final response after $K$ iterations.

\subsubsection{Early Stopping for Parallel Decoding}
\label{sec:early_stopping_dllm}

Diffusion-based LLMs (dLLMs) generate text by iteratively denoising a masked sequence.
Although additional steps can refine quality, many tokens reach their final values well before the last denoising step, so continuing refinement wastes compute and increases latency.
Early stopping accelerates inference by committing tokens once further updates are unlikely to change the final outcome.

Existing early-stopping methods are dominated by \emph{heuristic or step-local} controllers.
Heuristic rules (e.g., Fast-dLLM~\cite{wu2025fastdllmtrainingfreeaccelerationdiffusion}, Prophet~\cite{li2025diffusion}, Credit Decoding~\cite{wang2025creditdecoding} apply global thresholds on current-step statistics such as top-1 confidence or entropy, while learned controllers (e.g., Learn2PD~\cite{bao2025learn2pd}) train a lightweight predictor but still make thresholded accept/reject decisions at inference.
We show that this step-local, uniformly-applied design is unreliable due to the temporal and spatial structure of diffusion trajectories, and that passive waiting misses forecastable acceleration opportunities.

\textbf{Step-local and Spatial-uniform Control Is Unreliable.}
The same current-step confidence can correspond to very different future outcomes, and the ambiguity varies across token positions.
We validate this on a held-out C4~\cite{raffel2020exploring} subset by extracting token-states from intermediate denoising steps.
For position $i$ at step $k$, let $p_i^{(k)}\in\Delta^V$ be the predictive distribution; we group token-states with nearly indistinguishable step-local distributions (matched by entropy and tight Jensen–Shannon divergence) but different future trajectories.
We define the flip probability $P_{\text{flip}}(k)=\Pr[\hat{y}_{0,i}^{(K)}\neq \hat{y}_{0,i}^{(k)}]$ and observe that, within the same confidence bin, unstable traces have multiple-$\times$ higher flip rates than stable ones (Fig.~\ref{fig:temp}).
Moreover, stabilization is position-dependent: tokens at more rightward positions stabilize later (Fig.~\ref{fig:spatial}), so no single global threshold (heuristic or learned) can be well-calibrated for all positions simultaneously.

\textbf{Heuristic Controllers Are Fragile.}
Most prior methods ultimately threshold a step-local score with a fixed $\tau$.
This includes heuristic rules and learned filters such as Learn2PD~\cite{bao2025learn2pd}, which replace hand-crafted scores with a learned score $f_\theta(\cdot)$ but still decide ``accept'' if $f_\theta(\cdot)>\tau$.
When the score--stability relationship shifts across steps or positions, a static $\tau$ becomes either overly conservative (hurting speed) or overly aggressive (hurting quality), making tuning sensitive and workload-dependent.

\textbf{Passive Waiting Misses forecastable Acceleration.}
When uncertain, existing controllers typically wait for additional denoising steps until a threshold is crossed.
However, once a token enters a trend-consistent regime, its future confidence is often predictable from a short prefix.
Fig.~\ref{fig:extrap1} shows a case where the winning token follows a stable upward trend long before reaching a standard confidence threshold, suggesting that a look-ahead mechanism can commit earlier.
Fig.~\ref{fig:extrap2} further indicates that many tokens exhibit trend-consistent portions, leaving substantial headroom for saving steps.

These findings motivate two principles: (i) early stopping should be \emph{temporal- and position-aware}, leveraging token-wise trajectories and spatial heterogeneity rather than step-local uniform thresholds; and (ii) it should be \emph{proactive}, using predictable dynamics with calibrated uncertainty to anticipate stabilization instead of passively waiting.
As shown in Fig. \ref{caption}, we instantiate these principles with (i) TSPD, a trajectory-based controller that models token history and relative position, and (ii) a training-free confidence extrapolation module that forecasts near-future confidences with uncertainty to enable risk-limited look-ahead decisions.

\subsection{Temporal-Spatial Parallel Decoding (TSPD)}
\label{sec:tspd}

In practice, token-wise uncertainty evolves with diverse trajectories (e.g., monotone convergence, oscillation, or delayed stabilization), and these patterns are not reliably inferred from a single-step snapshot.
We therefore develop \emph{Temporal-Spatial Parallel Decoding (TSPD)}, a lightweight learnable controller that tracks per-token temporal traces and spatial position within each decoding block, and outputs a binary \emph{fix-or-continue} decision for every token at every diffusion step.

\textbf{Temporal-spatial Decoding Signals.}
Let $x^{(t)}=(x^{(t)}_1,\ldots,x^{(t)}_L)$ denote the intermediate sequence at diffusion step $t$ (with $t$ decreasing from $T$ to $0$), and let $p_\theta(\cdot \mid c, x^{(t)})$ be the dLLM token distribution.
For each position $i$, we take the top-1 candidate $\hat{x}^{(t)}_i=\arg\max_v p^{(t)}_i(v)$ and its confidence $p^{(t)}_i=\max_v p^{(t)}_i(v)$, and measure uncertainty by the predictive entropy $H^{(t)}_i=-\sum_v p^{(t)}_i(v)\log p^{(t)}_i(v)$.
To reduce step-wise noise, we track an exponential moving average confidence $\bar{p}^{(t)}_i=\alpha p^{(t)}_i+(1-\alpha)\bar{p}^{(t+1)}_i$ and use its one-step difference as a momentum feature $\Delta\bar{p}^{(t)}_i=\bar{p}^{(t)}_i-\bar{p}^{(t+1)}_i$.
We also include the token’s relative position within the block, $\phi(i)=i/(L-1)$.
Thus, at each step $t$ TSPD constructs the per-token feature vector as:
\begin{equation}
r^{(t)}_i \triangleq \big[p^{(t)}_i,\; H^{(t)}_i,\; \Delta \bar{p}^{(t)}_i,\; \phi(i)\big],
\label{eq:tspd_trace}
\end{equation}
which captures instantaneous evidence ($p^{(t)}_i$, $H^{(t)}_i$), temporal trend ($\Delta \bar{p}^{(t)}_i$), and spatial context ($\phi(i)$).

\textbf{Sequential binary controller.}
TSPD employs a shared recurrent controller to maintain a hidden state for each token position, enabling memory over the denoising trajectory.
Specifically, for each token $i$ we update
\begin{equation}
h^{(t)}_i = f_\psi\!\left(h^{(t+1)}_i,\; r^{(t)}_i\right),
\label{eq:tspd_rnn}
\end{equation}
where $f_\psi(\cdot)$ is a lightweight recurrent update (e.g., a GRU~\cite{cho2014learning}/LSTM~\cite{hochreiter1997long} cell) with parameters $\psi$.
A linear head produces a gate logit
\begin{equation}
z^{(t)}_i = W h^{(t)}_i + b,
\label{eq:tspd_logit}
\end{equation}
which is converted to a soft fixing propensity $\pi^{(t)}_i=\sigma(z^{(t)}_i)$.
Unlike threshold-tuned score predictors, TSPD executes a \emph{direct binary action} by hard gating:
\begin{equation}
a^{(t)}_i \in \{0,1\}, \qquad a^{(t)}_i = \mathbb{I}\big[\pi^{(t)}_i \ge 0.5\big],
\label{eq:tspd_binary}
\end{equation}
where $a^{(t)}_i=1$ indicates that token $i$ is \emph{fixed} (committed and excluded from further denoising) and $a^{(t)}_i=0$ keeps it active.

\textbf{Training by Oracle Imitation with Discrete Decisions.}
We supervise TSPD using an oracle fixing policy that marks a token as fixable once its candidate token matches the final target at that step.
Formally, let $y^{(t)}_i\in\{0,1\}$ denote the oracle \emph{fix} label for token $i$ at step $t$.
We train the controller via binary cross-entropy on logits:
\begin{equation}
\mathcal{L}_{\text{gate}} = \frac{1}{LT}\sum_{t=0}^{T}\sum_{i=1}^{L} \mathrm{BCEWithLogits}\!\left(z^{(t)}_i,\; y^{(t)}_i\right).
\label{eq:tspd_loss}
\end{equation}
To align training with inference-time hard gating, we use a straight-through estimator (STE): the forward pass applies the hard decision $a^{(t)}_i$ to simulate fixing, while gradients are backpropagated through the continuous $\pi^{(t)}_i$ (equivalently, through $z^{(t)}_i$ in \eqref{eq:tspd_loss}).
By jointly leveraging temporal traces and spatial position, TSPD provides robust parallel decoding control with negligible overhead.

\begin{algorithm}[t]
\caption{Inference Workflow}
\label{alg:tspd_ce_infer}
\begin{algorithmic}[1]
\REQUIRE dLLM $\mathcal{M}$ with $K$ steps; prompt $x$; generation length $L$; controller \textsc{TSPD}; extrapolator \textsc{CE}.
\STATE $X \gets \mathrm{concat}(x,\,[\texttt{MASK}]^{L})$; \;\; $F \gets \mathbf{0}$ \COMMENT{$F_j{=}1$ means token $j$ is fixed}
\STATE Initialize per-token trace buffer $\mathcal{H}$
\FOR{$t = K, \ldots, 1$}
    \STATE $p(\cdot) \gets \mathcal{M}(X)$ \COMMENT{one dLLM forward pass}
    \STATE Update trace history $\mathcal{H} \gets \text{update}(\mathcal{H}, p)$
    \STATE $\hat{r} \gets \textsc{CE}(\mathcal{H}, F)$
    \STATE $A \gets \textsc{TSPD}(\hat{r}, \text{pos}, F)$ \COMMENT{binary fix decisions}
    \STATE $X \gets \text{commit}(X, p, A)$; \;\; $F \gets F \lor A$
    \IF{$\sum F = L$} \STATE \textbf{break} \ENDIF
    \STATE $X \gets \text{remask}(X, p, F)$
\ENDFOR
\STATE \textbf{return} $\text{resp}(X)$
\end{algorithmic}
\end{algorithm}

\subsection{Confidence Extrapolation}
\label{sec:logit_extrapolation}

Most existing decoding accelerations are passive: when confidence is low, the model simply runs more denoising steps until the distribution sharpens.
We instead propose \emph{Confidence Extrapolation} (CE), a training-free, plug-in module that \emph{forecasts near-future confidence} from the observed confidence trace and feeds a calibrated, risk-aware confidence signal to any downstream decoding controller (e.g., TSPD, Learn2PD, or threshold rules).
CE does not modify the base diffusion model parameters and can be combined with existing controllers without changing their decision logic.

\textbf{State-space Extrapolator on Confidence.}
For each position $j$ in a decoding block, let $c_t(j)\in[0,1]$ denote the confidence at diffusion step $t$ (e.g., top-1 probability or a margin-based confidence).
To obtain a stable time series, we apply a monotone transform $y_t(j)=\phi(c_t(j))$ (e.g., log-odds).
We model $y_t(j)$ as a noisy observation of a latent confidence level:
\begin{equation}
y_t(j)=\ell_t(j)+n_t,\quad n_t\sim\mathcal{N}(0,R_t(j)).
\end{equation}
CE maintains a level--velocity latent state
\begin{equation}
x_t(j)=\big[\ell_t(j),\,\dot{\ell}_t(j)\big]^\top,\quad
x_{t+1}(j)=Ax_t(j)+w_t,
\label{eq:dynamics}
\end{equation}

where $Q$ determines how rapidly the underlying trend is allowed to change, and $A$ is a fixed two-by-two transition matrix with ones on the diagonal and the upper-right entry, and zeros elsewhere.
Given the filtered estimate $(\hat{x}_t,P_t)$, the $h$-step forecast is
\begin{equation}
\hat{x}_{t+h|t}=A^h\hat{x}_t,\qquad
P_{t+h|t}=A^hP_t(A^h)^\top+Q_h,
\label{eq:multi_step_pred}
\end{equation}
where $Q_h \triangleq \sum_{i=0}^{h-1} A^i Q (A^i)^\top$ is the accumulated process noise over $h$ steps.
And the forecast variance of the confidence level is $\sigma^2_{t+h}(j)=\big(P_{t+h|t}(j)\big)_{1,1}$.
We map forecasts back to confidence by $\hat{c}_{t+h}(j)=\phi^{-1}(\hat{\ell}_{t+h}(j))$.

\textbf{Risk-aware Horizon Selection.}
CE selects the \emph{furthest} look-ahead horizon that is both (i) supported by a reliable history window and (ii) safe under extrapolation uncertainty.
Although denoising within a block is parallel, later positions become reliable only after sufficient left context is fixed.
\noindent We use $[x]_0^1 \triangleq \min\{1,\max\{0,x\}\}$ to clamp $x$ into $[0,1]$.
Let $\textit{left\_coverage}_t(j)\in[0,1]$ be the fraction of fixed positions left of $j$ at step $t$, and define an extrapolation potential
\begin{equation}
r_t(j)=\left[\frac{\textit{left\_cvg}_t(j)-\tau}{1-\tau}\right]_0^1,\quad
h_{\text{pot}}= Hr_t(j).
\label{eq:hpot}
\end{equation}
where $\tau$ is a coverage threshold and $H$ is the maximum look-ahead.
For a candidate horizon $h\le h_{\mathrm{pot}}$, we form a conservative lower bound in the transformed space:
\begin{equation}
\mathrm{LB}_{t+h}\triangleq \hat{\ell}_{t+h}-z\,\sigma_{t+h},\qquad
\hat{c}^{\mathrm{LB}}_{t+h}=\phi^{-1}\!\big(\mathrm{LB}_{t+h}\big).
\label{eq:lower_bound}
\end{equation}
where $z$ controls risk tolerance. For brevity, we write $\hat{\ell}_{t+h}\equiv \hat{\ell}_{t+h\mid t}$ and omit the position index $j$ when clear.
We choose
\begin{equation}
h^{(t,j)}=\max\{\,h\le h_{\mathrm{pot}}(t,j):\ \hat{c}^{\mathrm{LB}}_{t+h}(j)\ge c_{\min}\,\}.
\label{eq:hstar}
\end{equation}
and fall back to $h^{(t,j)}=0$ if no horizon passes.
The controller then consumes $\tilde{c}_{t(j)}=\hat{c}^{\mathrm{LB}}_{t+h^{(t,j)}}(j)$ in place of the raw $c_{t(j)}$.

\textbf{Integration with Decoding Controllers.}
CE exposes a minimal interface that outputs $\tilde{c}_{t(j)}$ (and optionally $h^{(t,j)}$) per step and position, making it directly compatible with threshold-based or learned controllers.
As shown in Alg. \ref{alg:tspd_ce_infer}, we simply replace the controller's per-step confidence feature with $\tilde{c}_t(j)$, enabling earlier commitment when the forecast indicates a reliably increasing confidence trajectory.

\begin{table}[t]
\centering
\small
\caption{Benchmark results on the LLaDA-8B-Instruct suite with a generation length of 256. Performance is measured using TPS, speedup, and accuracy (\footnotesize $\dagger$ denotes without KV cache).}
\label{tab:llada_benchmark}

\begin{tabular}{@{} M{1.2cm} | M{1.8cm} | C{0.9cm} | C{1.1cm} | C{0.8cm} @{}}
\toprule
\multirow[c]{2}{*}{\textbf{Task}} &
\multirow[c]{2}{*}{\textbf{Methods}} &
\multicolumn{2}{c|}{\textbf{Inference Efficiency}} &
\multirow[c]{2}{*}{\textbf{Acc.}} \\
\cline{3-4}
& & \textbf{TPS$\uparrow$} & \textbf{Speed$\uparrow$} & \\
\hline

\multirow[c]{7}{*}{\makecell[c]{GSM8K\\(5-shot)}} &
Vanilla & 6.9 & 1.0$\times$ & 79.3 \\
\cline{2-5}
& F-dLLM & 53.8 & 7.8$\times$ & 78.5 \\
\cline{2-5}
& CD & 54.5 & 7.9$\times$ & 78.7 \\
\cline{2-5}
& Prophet$^\dagger$ & 11.0 & 1.6$\times$ & \textbf{79.4} \\
\cline{2-5}
& Learn2PD$^\dagger$ & 29.0 & 4.2$\times$ & 79.1 \\
\cline{2-5}
& Ours$^\dagger$ & 34.2 & 5.0$\times$ & \textbf{79.4} \\
\cline{2-5}
& Ours & \textbf{77.3} & \textbf{11.2$\times$} & 78.8 \\
\hline

\multirow[c]{7}{*}{\makecell[c]{MATH\\(4-shot)}} &
Vanilla & 9.2 & 1.0$\times$ & \textbf{33.5} \\
\cline{2-5}
& F-dLLM & 51.5 & 5.6$\times$ & 33.2 \\
\cline{2-5}
& CD & 50.6 & 5.5$\times$ & 33.1 \\
\cline{2-5}
& Prophet$^\dagger$ & 15.7 & 1.7$\times$ & 32.7 \\
\cline{2-5}
& Learn2PD$^\dagger$ & 29.4 & 3.2$\times$ & 32.3 \\
\cline{2-5}
& Ours $^\dagger$ & 36.8 & 4.0$\times$ & 33.0 \\
\cline{2-5}
& Ours & \textbf{76.4} & \textbf{8.3$\times$} & 33.2 \\
\hline

\multirow[c]{7}{*}{\makecell[c]{HumanEval\\(0-shot)}} &
Vanilla & 6.5 & 1.0$\times$ & 41.5 \\
\cline{2-5}
& F-dLLM & 24.1 & 3.7$\times$ & 43.3 \\
\cline{2-5}
& CD & 24.7 & 3.8$\times$ & \textbf{43.7} \\
\cline{2-5}
& Prophet $^\dagger$ & 7.8 & 1.2$\times$ & 41.5 \\
\cline{2-5}
& Learn2PD $^\dagger$ & 23.4 & 3.6$\times$ & 40.3 \\
\cline{2-5}
& Ours $^\dagger$ & 27.3 & 4.2$\times$ & 42.5 \\
\cline{2-5}
& Ours & \textbf{53.3} & \textbf{8.2$\times$} & 43.0 \\
\hline

\multirow[c]{7}{*}{\makecell[c]{MBPP\\(3-shot)}} &
Vanilla & 5.8 & 1.0$\times$ & 29.4 \\
\cline{2-5}
& F-dLLM & 41.8 & 7.2$\times$ & 28.2 \\
\cline{2-5}
& CD & 40.6 & 7.0$\times$ & 29.3 \\
\cline{2-5}
& Prophet $^\dagger$ & 8.1 & 1.4$\times$ & 29.2 \\
\cline{2-5}
& L2PD $^\dagger$ & 29.6 & 5.1$\times$ & 29.2 \\
\cline{2-5}
& Ours $^\dagger$ & 32.5 & 5.6$\times$ & \textbf{29.5} \\
\cline{2-5}
& Ours & \textbf{68.4} & \textbf{11.8$\times$} & 29.2 \\
\bottomrule
\end{tabular}
\end{table}

\begin{table}[t]
\centering
\small
\caption{Benchmark results on the Dream-7B and LLaDA-MoE suite with a generation length of 256. Performance is measured using TPS (tokens/sec), speedup, and accuracy
(\footnotesize $\dagger$ denotes without KV cache).}
\label{tab:llada_dream_benchmark}

\begin{tabular}{@{} M{1.8cm} | M{1.6cm} | C{0.9cm} | C{1.1cm} | C{0.8cm} @{}}
\toprule
\multirow[c]{1}{*}{\textbf{Task}} &
\multirow[c]{1}{*}{\textbf{Methods}} &
\multirow[c]{1}{*}{\textbf{TPS$\uparrow$}} &
\multirow[c]{1}{*}{\textbf{Speed$\uparrow$}} &
\multirow[c]{1}{*}{\textbf{Acc.}} \\
\hline

\multirow[c]{7}{*}{\makecell[c]{Dream-7B}} &
Vanilla & 9.1 & 1.0$\times$ & 75.0 \\
\cline{2-5}
& F-dLLM & 48.2 & 5.3$\times$ & 74.8 \\
\cline{2-5}
& CD & 42.8 & 4.7$\times$ & 74.5 \\
\cline{2-5}
& Prophet$^\dagger$ & 15.1 & 1.7$\times$ & 74.9 \\
\cline{2-5}
& Learn2PD$^\dagger$ & 19.1 & 2.1$\times$ & 73.4 \\
\cline{2-5}
& Ours$^\dagger$ & 28.2 & 3.1$\times$ & \textbf{75.3} \\
\cline{2-5}
& Ours & \textbf{69.2} & \textbf{7.6$\times$} & 75.0 \\
\hline

\multirow[c]{7}{*}{\makecell[c]{LLaDA-MoE}} &
Vanilla & 4.2 & 1.0$\times$ & 75.2 \\
\cline{2-5}
& F-dLLM & 9.7 & 2.3$\times$ & 75.3 \\
\cline{2-5}
& CD & 10.5 & 2.5$\times$ & 75.2 \\
\cline{2-5}
& Prophet$^\dagger$ & 6.7 & 1.6$\times$ & 75.0 \\
\cline{2-5}
& Learn2PD$^\dagger$ & 13.0 & 3.1$\times$ & 75.3 \\
\cline{2-5}
& Ours$^\dagger$ & 13.8 & 3.3$\times$ & \textbf{75.4} \\
\cline{2-5}
& Ours & \textbf{21.2} & \textbf{5.1$\times$} & 75.2 \\
\bottomrule
\end{tabular}
\end{table}

\section{Experiments}

\subsection{Experimental Settings}
\label{sec:exp_settings}

\textbf{Models and Datasets.}
We mainly implement our method on the representative dLLM: LLaDA-8B-Instruct~\cite{nie2025large} to measure the acceleration of the inference process across various benchmarks.
To ensure the broad applicability of the method, we conduct experiments on four datasets covering three different types of problems, which are GSM8K~\cite{cobbe2021training}, MATH~\cite{hendrycks2021measuring}, HumanEval~\cite{chen2021evaluating}, and MBPP~\cite{austin2021program}. To demonstrate the generalizability of our method across diverse dLLM architectures, we conduct additional comparisons with existing methods on Dream-7B~\cite{ye2025dream7b} and LLaDA-MoE~\cite{zhu2025llada}, further confirming the effectiveness of our approach. In this paper, we primarily compare our method with state-of-the-art dLLM acceleration approaches, including Fast-dLLM~\cite{wu2025fastdllmtrainingfreeaccelerationdiffusion}, Credit Decoding (CD)~\cite{wang2025creditdecoding},  Prophet~\cite{li2025diffusion}, and Learn2PD~\cite{bao2025learn2pd}. Among these methods, Fast-dLLM and Credit Decoding leverage Key-Value (KV) caching, whereas Learn2PD and Prophet do not utilize KV cache mechanisms.

\textbf{TSPD Training.}
To train the TSPD model that can be applied to a wide range of tasks, we select 40 samples from each of the 66 types of questions in the FLAN dataset (out of the used evaluation benchmarks), resulting in a total of 2,640 samples for training.
In this experiment, we adopt the simplest two-layer LSTM as our filter model.
Since the dLLMs remain frozen and only $f_\theta$ is trained, the number of trainable parameters is extremely limited.
For example, for an LLaDA with a block size of 32, the total number of trainable parameters is only 2,112.
We train the $f_\theta$ for 5,000 epochs until the model converged.
The learning rate is set to 0.001, and the AdamW optimizer is used to optimize $f_\theta$.

Our training process consists of two stages.
In the first stage, samples are collected by following an Extremely Greedy Parallel~\cite{bao2025learn2pd} policy, recording the confidence scores and token selections at each step during parallel decoding.
This data is then used in the second stage to train the TSPD model $f_\theta$.
The data collection in the first stage is conducted on 2 NVIDIA RTX A100 GPUs and takes approximately two hours.
The subsequent training of the TSPD model in the second stage is deployed on the same GPU and required only 6 minutes.
The details of training are provided in Appendix~\ref{app:stage2}.

\textbf{Evaluation.}
We evaluate the inference acceleration and generation quality of TSPD and CE methods using quantitative metrics.
The inference speed is quantified with Tokens Per Second (TPS), indicating the average number of tokens generated per second.
The generation quality is measured using task-specific metrics, such as accuracy for GSM8K, reflecting the model’s performance under acceleration.
In addition, we set the generation length to 256 and 1024, and the block size to 32.

\begin{table}[t]
\centering
\small
\caption{Benchmark results on the LLaDA-8B-Instruct suite with a generation length of 512 and 1024. \footnotesize $\dagger$ denotes without KV cache.}
\label{tab:llada_variable_len_benchmark}

\begin{tabular}{@{} M{1.2cm} | M{2.4cm} | C{0.8cm} | C{1.0cm} | C{0.8cm} @{}}
\toprule
\multirow[c]{1}{*}{\textbf{Gen Len.}} &
\multirow[c]{1}{*}{\textbf{Methods}} &
\multirow[c]{1}{*}{\textbf{TPS$\uparrow$}} &
\multirow[c]{1}{*}{\textbf{Speed$\uparrow$}} &
\multirow[c]{1}{*}{\textbf{Acc.}} \\
\hline

\multirow[c]{5}{*}{\makecell[c]{512}} &
Vanilla & 3.2 & 1.0$\times$ & 77.5 \\
\cline{2-5}
& Fast-dLLM       & 35.3 & 11.0$\times$ & 77.2 \\
\cline{2-5}
& Learn2PD$^\dagger$        & 39.4 & 12.3$\times$ & 77.8 \\
\cline{2-5}
& Ours$^\dagger$ & 45.4 & 14.2$\times$ & \textbf{77.9} \\
\cline{2-5}
& Ours & \textbf{79.0} & \textbf{24.7$\times$} & 77.6 \\
\hline

\multirow[c]{5}{*}{\makecell[c]{1024}} &
Vanilla & 1.1 & 1.0$\times$ & 77.0 \\
\cline{2-5}
& Fast-dLLM       & 21.6 & 19.6$\times$ & 74.7 \\
\cline{2-5}
& Learn2PD$^\dagger$     & 24.9 & 22.6$\times$ & 78.7 \\
\cline{2-5}
& Ours$^\dagger$ & 27.3 & 24.8$\times$ & \textbf{79.3} \\
\cline{2-5}
& Ours & \textbf{64.1} & \textbf{58.3$\times$} & 78.5 \\
\bottomrule
\end{tabular}
\end{table}

\subsection{Main Results}
\label{sec:main_results}

\textbf{Overall Performance.}
Table~\ref{tab:llada_benchmark} reports results on the LLaDA-8B-Instruct suite with generation length 256.
Across GSM8K, MATH, HumanEval, and MBPP, our method consistently achieves the highest throughput and speedup while preserving accuracy close to Vanilla.
With KV cache, our method reaches \textbf{77.3} TPS / \textbf{11.2$\times$} on GSM8K, \textbf{76.4} / \textbf{8.3$\times$} on MATH, \textbf{53.3} / \textbf{8.2$\times$} on HumanEval, and \textbf{68.4} / \textbf{11.8$\times$} on MBPP.
Without cache, our method still outperforms Prophet and Learn2PD with same settings on all tasks (e.g., 34.2 vs.\ 29.0 TPS on GSM8K, 36.8 vs.\ 29.4 on MATH), demonstrating that our gains do not rely on additional caching.

\textbf{Performance over Different dLLMs.}
Table~\ref{tab:llada_dream_benchmark} shows that the improvements transfer across backbones.
On Dream-7B, our method achieves \textbf{69.2} TPS / \textbf{7.6$\times$} with accuracy matching Vanilla (75.0\%), and our method (without KV cache) remains competitive (28.2 TPS / 3.1$\times$).
On LLaDA-MoE, our method improves throughput to \textbf{21.2} TPS / \textbf{5.1$\times$} with essentially unchanged accuracy (75.2\%), indicating that CE and the controller are not model-specific and generalize to MoE architectures.

\begin{table}[t]
\centering
\small
\caption{A comparison of our method with and without KV Cache. The results show a significant performance improvement when augmented with both Dual and Prefix Caches, underscoring that our method is orthogonal to and fully compatible with existing KV caching strategies.}
\label{tab:kv_cache_ablation}
\setlength{\tabcolsep}{10pt}
\begin{tabular}{l|c|c|c}
\toprule
\textbf{Methods} & \textbf{TPS$\uparrow$} & \textbf{Speed$\uparrow$} & \textbf{Acc.} \\
\midrule
TSPD \& CE & 34.2 & 5.0$\times$ & \textbf{79.4} \\
\midrule
\quad + Dual Cache   & \textbf{77.3} & \textbf{11.2$\times$} & 78.8 \\
\quad + Prefix Cache & 37.3 & 5.4$\times$ & 79.3 \\
\bottomrule
\end{tabular}
\end{table}

\begin{table}[t]
\centering
\small
\caption{Ablation of the effectiveness of TSPD and CE modules using LLaDA-8B-Instruct model on GSM8K benchmark at generation length of 256 tokens. \footnotesize $\dagger$ denotes without KV cache.}
\label{tab:module_ablation}
\setlength{\tabcolsep}{10pt}
\begin{tabular}{l|c|c|c}
\toprule
\textbf{Methods} & \textbf{TPS$\uparrow$} & \textbf{Speed$\uparrow$} & \textbf{Acc.} \\
\midrule
Ours$^\dagger$  & \textbf{34.2} & \textbf{5.0$\times$} & 79.4 \\
\midrule
w.o. TSPD  & 20.0 & 2.9$\times$ & 79.1 \\
w.o. CE & 30.3 & 4.4$\times$ & \textbf{79.5} \\
\bottomrule
\end{tabular}
\end{table}

\begin{table}[t]
\centering
\small
\caption{A comparison of the acceleration performance using filter models of varying complexity. The results indicate that a two-layer LSTM model achieves the optimal balance by providing significant speedup and maintaining high accuracy.}
\label{tab:arch_ablation}
\setlength{\tabcolsep}{10pt}
\begin{tabular}{l|c|c|c}
\toprule
\textbf{\# Layers} & \textbf{TPS$\uparrow$} & \textbf{Speed$\uparrow$} & \textbf{Acc.} \\
\midrule
1-layer MLP & 29.7  & 4.3$\times$ & 78.7 \\
2-layer MLP & \textbf{30.4} & \textbf{4.4$\times$} & 78.8 \\
4-layer MLP & 26.8 & 3.9$\times$ & 79.0 \\
1-layer LSTM & 29.0 & 4.2$\times$ & 79.2 \\
2-layer LSTM & 30.3 & \textbf{4.4$\times$} & \textbf{79.5} \\
4-layer LSTM & 28.3 & 4.1$\times$ & \textbf{79.5} \\
\bottomrule
\end{tabular}
\end{table}

\textbf{Performance over Different Generation Lengths.}
Table~\ref{tab:llada_variable_len_benchmark} evaluates longer generations (512/1024).
Speedups increase with length: at 512 tokens, our method achieves \textbf{79.0} TPS / \textbf{24.7$\times$}; at 1024 tokens, it reaches \textbf{64.1} TPS / \textbf{58.3$\times$}, while maintaining accuracy comparable to Vanilla.
In contrast, Fast-dLLM shows a larger accuracy drop at 1024 (from 77.0\% to 74.7\%).
Overall, our method scales favorably with generation length, yielding larger end-to-end acceleration without sacrificing generation quality.

\subsection{Compatibility with Key-Value Cache}
\label{sec:kv_compatibility}

We evaluate compatibility with KV caching~\cite{pope2023efficiently,liu2024minicache} by integrating Dual Cache~\cite{wu2025fastdllmtrainingfreeaccelerationdiffusion} and Prefix Cache~\cite{wu2025fastdllmtrainingfreeaccelerationdiffusion} on GSM8K (256 tokens).
As shown in Table~\ref{tab:kv_cache_ablation}, the baseline TSPD\,+\,CE achieves 34.2 TPS (5.0$\times$) with 79.4\% accuracy.
Adding Dual Cache boosts throughput to 77.3 TPS (11.2$\times$) with a modest accuracy drop (78.8\%), while Prefix Cache improves to 37.3 TPS (5.4$\times$) with 79.3\% accuracy.
Overall, our method is orthogonal to standard KV caching and can directly benefit from these techniques for further speedups.

\begin{table}[t]
\centering
\small
\caption{A ablation of the effectiveness of the features used for TSPD using LLaDA-8B-Instruct model on GSM8K benchmark at generation length of 256 tokens.}
\label{tab:feat_ablation}
\setlength{\tabcolsep}{10pt}
\begin{tabular}{l|c|c|c}
\toprule
\textbf{Methods} & \textbf{TPS$\uparrow$} & \textbf{Speed$\uparrow$} & \textbf{Acc.} \\
\midrule
TSPD & 30.3 & 4.4$\times$ & \textbf{79.5} \\
\midrule
w.o. confidence  & 19.3 & 2.8$\times$ & 78.8 \\
w.o. entropy & 28.3 & 4.1$\times$ & 79.0 \\
w.o. momentum & 29.7 & 4.3$\times$ & 79.2 \\
w.o. position & \textbf{31.1} & \textbf{4.5$\times$ }& 78.5 \\
\bottomrule
\end{tabular}
\end{table}

\subsection{Sensitivity Analysis}

\textbf{Parameter Budget on TSPD.}
We vary the TSPD parameter budget within the same 2-layer LSTM architecture by scaling the hidden size. As shown in Table~\ref{tab:tspd_param_scale}, the current \textbf{2K} setting provides the best speed--accuracy tradeoff. Reducing the budget to \textbf{1K} slightly improves TPS (\(30.9\rightarrow 30.3\)) but lowers accuracy (\(79.2\rightarrow 79.5\)). Increasing the budget to \textbf{4K} or \textbf{8K} brings at most marginal accuracy gain, while throughput drops more noticeably. This suggests that the current controller is already near the sweet spot: smaller models underfit the temporal traces, whereas larger ones add complexity with little benefit.

\begin{table}[t]
\centering
\caption{Impact of parameter budget on TSPD performance under the same 2-layer LSTM architecture.}
\label{tab:tspd_param_scale}
\begin{tabular}{lccc}
\toprule
Param budget & Hidden size & TPS$\uparrow$ & Acc.\ $\uparrow$ \\
\midrule
1K  & 8  & \textbf{30.9} & 79.2 \\
2K  & 12 & 30.3 & \textbf{79.5} \\
4K  & 17 & 29.8 & 79.5 \\
8K  & 25 & 26.8 & 79.7 \\
\bottomrule
\end{tabular}
\end{table}

\textbf{Training Sensitivity to Data Collection Policy.}
We compare EGP-collected, standard parallel-decoding, and mixed training traces under the same test-time setting. As shown in Table~\ref{tab:train_policy_sensitivity}, EGP gives the best tradeoff (\textbf{30.3 TPS}, \textbf{79.5 Acc}), while standard traces are much slower (\textbf{7.5 TPS}) with similar accuracy; the mixed policy is in between (\textbf{16.1 TPS}, \textbf{79.5 Acc}).
This suggests that EGP does not cause harmful distribution shift, but instead provides more informative borderline states for learning when to safely fix tokens.

\begin{table}[t]
\centering
\caption{Training sensitivity of data collection policy.}
\label{tab:train_policy_sensitivity}
\begin{tabular}{lccc}
\toprule
 & EGP & Standard PD & Mixed \\
\midrule
TPS $\uparrow$ & \textbf{30.3} & 7.5 & 16.1 \\
Acc $\uparrow$ & 79.5 & \textbf{79.6} & 79.5 \\
\bottomrule
\end{tabular}
\end{table}

\textbf{Sensitivity to Training Data Scale.}

We evaluate TSPD with different numbers of FLAN training prompts while keeping the controller architecture and decoding settings fixed. As shown in Table~\ref{tab:tspd_data_scale}, TSPD is already close to convergence with \textbf{1320} samples: compared with the default \textbf{2640}-sample setting, the accuracy gap is only \textbf{0.1}, and further increasing the data to \textbf{3960} samples brings negligible gains. This is expected because TSPD is a lightweight controller trained on compact trajectory features rather than full token logits.

\begin{table}[t]

\centering

\caption{Sensitivity to TSPD training data scale.}

\label{tab:tspd_data_scale}

\begin{tabular}{lcc}

\toprule

Training Samples & TPS$\uparrow$ & Acc.\ $\uparrow$ \\

\midrule

Vanilla & 6.9 & 79.3 \\

330  & 27.1 & 78.8 \\

660  & 28.7 & 79.1 \\

1320 & 30.0 & 79.4 \\

2640 & \textbf{30.3} & \textbf{79.5} \\

3960 & 30.2 & \textbf{79.5} \\

\bottomrule

\end{tabular}

\end{table}

\textbf{Sensitivity to Training Data Distribution.}

We evaluate whether TSPD is sensitive to the difficulty distribution of training traces, measured by the average \emph{stabilization-step ratio}, (i.e., the fraction of denoising steps where a token already matches its final prediction). Lower ratios indicate harder traces with later stabilization, while higher ratios indicate easier traces with earlier stabilization. As shown in Table~\ref{tab:data_dist_sensitivity}, TSPD is stable across different distributions, with variations within \textbf{1.1 TPS} and \textbf{0.5 Acc}. Harder traces make the controller slightly more conservative and improve accuracy, while easier traces make it more aggressive and improve TPS. Our default mixed distribution (\(\sim0.45\)) gives the best overall tradeoff.

\begin{table}[t]

\centering

\caption{Sensitivity to training data distribution.}

\label{tab:data_dist_sensitivity}

\begin{tabular}{lcc}

\toprule

Stabilization Ratio & TPS$\uparrow$ & Acc.\ $\uparrow$ \\

\midrule

Vanilla & 6.9 & 79.3 \\

$\sim0.25$ & 29.6 & \textbf{79.7} \\

$\sim0.45$ & 30.3 & 79.5 \\

$\sim0.50$ & 30.1 & 79.5 \\

$\sim0.75$ & \textbf{30.7} & 79.2 \\

\bottomrule

\end{tabular}

\end{table}

\textbf{Sensitivity to CE Hyper-parameters.}
We evaluate the two key CE hyper-parameters, the coverage threshold $\tau$ and the maximum look-ahead horizon $H$, while keeping all other settings fixed. The masking ratio $\alpha$ belongs to the base dLLM denoising schedule rather than our method, and is therefore not tuned here. As shown in Table~\ref{tab:ce_hyper_sensitivity}, CE shows a stable speed--accuracy tradeoff: smaller $\tau$ or larger $H$ makes CE more aggressive and improves TPS, while larger $\tau$ or smaller $H$ better preserves accuracy. We use $\tau=0.6$ and $H=20$ by default.

\begin{table}[t]
\centering
\caption{Sensitivity to CE hyper-parameters.}
\label{tab:ce_hyper_sensitivity}
\begin{minipage}{0.48\linewidth}
\centering
\subcaption{Coverage threshold $\tau$.}
\label{tab:tau_sensitivity}
\begin{tabular}{lcc}
\toprule
$\tau$ & TPS$\uparrow$ & Acc.\ $\uparrow$ \\
\midrule
Vanilla & 6.9 & \textbf{79.3} \\
0.4 & \textbf{79.2} & 78.2 \\
0.5 & 77.0 & 78.7 \\
0.6 & 77.3 & 78.8 \\
0.7 & 75.2 & 78.8 \\
0.8 & 74.9 & 78.9 \\
\bottomrule
\end{tabular}
\end{minipage}
\hfill
\begin{minipage}{0.48\linewidth}
\centering
\subcaption{Max horizon $H$.}
\label{tab:H_sensitivity}
\begin{tabular}{lcc}
\toprule
$H$ & TPS$\uparrow$ & Acc.\ $\uparrow$ \\
\midrule
Vanilla & 6.9 & \textbf{79.3} \\
10 & 70.1 & 79.0 \\
15 & 72.4 & 78.8 \\
20 & 77.3 & 78.8 \\
25 & 78.0 & 78.2 \\
30 & \textbf{78.2} & 77.9 \\
\bottomrule
\end{tabular}
\end{minipage}
\end{table}

\subsection{Ablation Study}

\textbf{Effectiveness of TSPD and CE.} As shown in Tab. \ref{tab:module_ablation}, on GSM8K with generation length 256, both TSPD and CE improve throughput with negligible accuracy change.
Vanilla decoding runs at 6.9 TPS with 79.3\% accuracy.
Removing TSPD drops throughput to 20.0 TPS (2.9$\times$), indicating that trace-aware token fixing provides the main speedup.
Removing CE yields 30.3 TPS (4.4$\times$), showing that confidence extrapolation brings additional gains beyond TSPD alone.
Combining both achieves the best throughput, 34.2 TPS (5.0$\times$), while maintaining comparable accuracy (79.4\%).

\textbf{Complexity and Overhead Analysis.}
Table~\ref{tab:arch_ablation} shows that moderate-capacity filters perform best.
While a two-layer MLP achieves the highest efficiency among MLP variants (30.4 TPS, 4.4× speedup), a two-layer LSTM strikes a more favorable balance, delivering substantial speedup while maintaining high accuracy (30.3 TPS, 4.4× speedup, and 79.5\% accuracy). Since deeper variants yield diminishing returns,
we adopt the two-layer LSTM as the default.
TSPD and CE operate only on lightweight confidence features (per-position scalars and short traces), so their runtime overhead is negligible compared to a diffusion model forward.
Profiling on GSM8K (256 tokens) shows that the per-step latency of TSPD and CE is only a small fraction of one base-model forward by 0.30\% and 0.13\%, respecively.
Thus, the observed speedups primarily come from reducing redundant denoising steps rather than shifting cost to the controller.

\textbf{Ablation of the Features Used for TSPD.}
Table~\ref{tab:feat_ablation} ablates TSPD features, removing \emph{confidence} causes the largest drop (30.3$\rightarrow$19.3 TPS; 4.4$\times\rightarrow$2.8$\times$), confirming it as the key signal.
Entropy and momentum provide smaller but consistent gains (w.o.\ entropy: 4.1$\times$; w.o.\ momentum: 4.3$\times$).
Removing \emph{position} slightly improves speed (31.1 TPS, 4.5$\times$) but harms quality (78.5\% accuracy), indicating position mainly prevents premature fixing for later tokens.
Overall, confidence is essential; entropy/momentum are complementary; position improves accuracy.

\textbf{Confidence Extrapolation with Different Controllers.}
Table~\ref{tab:ce_ablation} shows CE consistently improves efficiency across controllers.
For TSPD, CE increases throughput from 30.3 to 34.2 TPS and speedup from 4.4$\times$ to 5.0$\times$ with similar quality; Learn2PD and heuristics used in Fast-dLLM exhibit similar gains.
This indicates CE is controller-agnostic and complements existing stopping policies by supplying forward-looking confidence.

\begin{table}[t]
\centering
\small
\caption{A comparison of our confidence extrapolator method with different decoding controllers using LLaDA-8B-Instruct model on GSM8K benchmark at generation length of 256 tokens.}
\label{tab:ce_ablation}
\setlength{\tabcolsep}{10pt}
\begin{tabular}{l|c|c|c}
\toprule
\textbf{Methods} & \textbf{TPS$\uparrow$} & \textbf{Speed$\uparrow$} & \textbf{Acc.} \\
\midrule
TSPD  & 30.3 & 4.4$\times$ & \textbf{79.5} \\
TSPD \& CE & 34.2 & 5.0$\times$ & 79.4 \\
\midrule
Learn2PD  & 29.0 & 4.2$\times$ & 79.1 \\
Learn2PD \& CE & 32.4 & 4.7$\times$ & 79.3 \\
\midrule
Fast-dLLM  & 52.8 & 7.8$\times$ & 78.5 \\
Fast-dLLM \& CE & \textbf{64.2} & \textbf{9.3$\times$} & 78.8 \\
\midrule
CD & 54.5 & 7.9$\times$ & 78.7 \\
CD \& CE & 58.0 & 8.4$\times$ & 78.7 \\
\bottomrule
\end{tabular}
\end{table}

\section{Conclusion}
\label{sec:conclusion}

In this work, we address dLLM inference inefficiency and unreliability caused by repetitive denoising and step-local stopping criteria. We propose Temporal-Spatial Parallel Decoding (TSPD), a lightweight trace-aware controller that uses token-wise trajectories and token position to decide when tokens have converged and can be fixed. We further introduce Confidence Extrapolation (CE), a training-free state-space plug-in that forecasts future confidences with uncertainty to enable proactive acceleration. Across four benchmarks on LLaDA-8B-Instruct (GSM8K, MATH, HumanEval, MBPP), TSPD delivers consistent and distinguished throughput gains at different generation settings with negligible accuracy change. Our approach also composes with KV caching for additional speedups, indicating strong compatibility with system optimizations.

\section*{Impact Statement}

This work enhances inference efficiency of diffusion-based large language models (dLLMs) via lightweight, training-free techniques (TSPD and Confidence Extrapolation), reducing latency and computation while preserving output quality. This facilitates practical deployment in resource-limited settings like edge devices, real-time applications, and large-scale systems.

Positive impacts include lower energy consumption for sustainable AI and greater accessibility to high-quality generation in education, scientific reasoning, coding, and productivity tools, democratizing advanced generative AI.

Faster inference may heighten misuse risks (e.g., rapid misinformation or harmful content). However, our method introduces no new vulnerabilities and is fully compatible with existing safeguards (alignment, filtering, watermarking). Responsible deployment practices should adequately mitigate concerns. Overall, the efficiency gains are predominantly beneficial for machine learning and real-world applications.

\bibliography{example_paper}
\bibliographystyle{icml2026}

\newpage
\appendix
\onecolumn

\section{Implementation and Training Details}
\label{app:impl_details}

\subsection{Software and Hardware}
\label{app:sw_hw}
We implement all methods in PyTorch (v2.1) with CUDA (v12.1) and run inference in FP16.
Experiments are executed on NVIDIA A100 80GB GPUs.
Unless noted otherwise, we use batch size $1$ and greedy decoding to avoid confounding factors from sampling.
Throughput is reported as \texttt{tokens/sec} (TPS) computed by
$
\mathrm{TPS} = \frac{B \cdot L_{\text{gen}}}{T_{\text{wall}}},
$
where $B$ is the batch size and $T_{\text{wall}}$ is the end-to-end wall-clock time from the first denoising step to termination, including remasking, controller overhead, and optional KV caching.
For timing stability, we warm up 20 runs and report the mean over 200 samples (or the full evaluation set if smaller).

\subsection{Base dLLM Decoding Configuration}
\label{app:base_decode}
We follow the default inference configuration of LLaDA-8B-Instruct.
The response span has length $L_{\text{gen}}{=}256$ and is initialized as all \texttt{[MASK]} tokens.
Denoising proceeds for at most $K{=}256$ steps.
At step $t$, the model outputs per-position logits $\ell_i^{(t)} \in \mathbb{R}^{V}$ and the token distribution
$
p_i^{(t)}(v) = \mathrm{softmax}(\ell_i^{(t)})_v .
$
We compute the top-1 prediction $\hat{y}_i^{(t)} = \arg\max_v p_i^{(t)}(v)$ and confidence
$
c_i^{(t)} = \max_v p_i^{(t)}(v),
$
and use the standard low-confidence remasking rule.
Let $\mathcal{A}^{(t)}$ be the set of \emph{active} (unfixed) positions at step $t$.
We remask the $n_t$ lowest-confidence positions in $\mathcal{A}^{(t)}$, where $n_t$ follows the default schedule used by LLaDA.
Concretely, we set
$
n_t = \left\lceil \rho(t)\cdot |\mathcal{A}^{(t)}| \right\rceil,\quad
\rho(t) = \rho_{\max}\cdot \Big(\frac{t}{K}\Big)^\gamma,
$
with $\rho_{\max}=0.5$ and $\gamma=1$ as a representative setting, and we keep fixed tokens excluded from remasking.
Decoding terminates when all response tokens are fixed or when $t=1$.

\subsection{TSPD Controller: Architecture and Features}
\label{app:tspd_arch}
TSPD is a per-token sequential classifier shared across positions.
We use a 2-layer LSTM with hidden size $d_h{=}16$ and input dimension $d_{\text{in}}{=}6$, followed by a linear head that outputs a Bernoulli logit.
For each token position $i$ at step $t$, we compute a compact trace feature vector
$
r_i^{(t)} = [\, c_i^{(t)},\ H_i^{(t)},\ \bar{c}_i^{(t)},\ \Delta\bar{c}_i^{(t)},\ \phi(i),\ u_i^{(t)} \,],
$
where entropy is
$
H_i^{(t)} = -\sum_{v} p_i^{(t)}(v)\log p_i^{(t)}(v),
$
EWMA confidence is
$
\bar{c}_i^{(t)} = \alpha c_i^{(t)} + (1-\alpha)\bar{c}_i^{(t+1)} \;\; \text{with}\;\; \alpha=0.25,
$
and momentum is $\Delta\bar{c}_i^{(t)}=\bar{c}_i^{(t)}-\bar{c}_i^{(t+1)}$.
$\phi(i)\in[0,1]$ denotes the relative position in the response span.
$u_i^{(t)}$ is an optional uncertainty feature provided by CE (Section~\ref{app:ce_impl}); when CE is disabled we set $u_i^{(t)}=0$.
TSPD updates a hidden state $h_i^{(t)}$ from $(h_i^{(t+1)}, r_i^{(t)})$ and outputs a fixing probability
$
\pi_i^{(t)}=\sigma(W h_i^{(t)}+b).
$
We fix token $i$ at step $t$ if $\pi_i^{(t)} \ge 0.5$, commit $\hat{y}_i^{(t)}$ to the sequence, and mark it as fixed for the remainder of decoding.

\subsection{Parameter Count and Overhead}
\label{app:tspd_overhead}
The controller is intentionally lightweight.
With $d_h{=}16$ and $d_{\text{in}}{=}6$, the 2-layer LSTM has approximately
$
\sum_{\ell=1}^{2}4(d_\ell(d_{\ell-1}+d_\ell)+d_\ell)
\approx 2{,}000
$
parameters (plus a small linear head), matching the order-of-magnitude reported in Section~\ref{sec:exp_settings}.
At inference, TSPD evaluates one LSTM step per active token per denoising step.
In practice this overhead is negligible relative to a full forward pass of the 8B dLLM; the controller runs on CPU or GPU and uses only vector operations on low-dimensional states.

\subsection{Training Data Collection (Stage 1)}
\label{app:stage1}
We collect supervision traces using an Extremely Greedy Parallel policy~\cite{bao2025learn2pd}.
For each prompt, we run the base dLLM to completion for $K$ steps and store, for each position $i$ and step $t$:
(i) trace features $r_i^{(t)}$,
(ii) the current top-1 token $\hat{y}_i^{(t)}$,
and (iii) the final token $\hat{y}_i^{(1)}$.
We then assign a binary label
$
y_i^{(t)} = \mathbb{I}\big[\hat{y}_i^{(t)}=\hat{y}_i^{(1)}\big],
$
which indicates whether fixing at step $t$ preserves the final outcome.
To reduce storage, we store only the feature tensors and labels (not full vocabulary logits).
For the FLAN subset used in Section~\ref{sec:exp_settings} (2,640 prompts) with $L_{\text{gen}}{=}256$ and $K{=}256$, this yields on the order of $2{,}640\times 256\times 256$ token-step instances before filtering; we subsample active positions and steps to control dataset size.

\subsection{Training Objective and Optimization (Stage 2)}
\label{app:stage2}
We train TSPD with weighted binary cross-entropy to address label imbalance:
$
\mathcal{L} = - w_1 y \log \pi - w_0 (1-y)\log(1-\pi),
$
with $w_1/w_0$ set by inverse class frequency on the training split.
We use AdamW with learning rate $10^{-3}$, weight decay $10^{-2}$.
We train for 5,000 epochs with early stopping on a held-out 10\% validation split.
To improve generalization across tasks, we randomly drop trace channels during training (dropout $p=0.1$ on $r_i^{(t)}$) and apply small Gaussian noise to confidence-related features ($\sigma=0.01$).
The dLLM remains frozen; only TSPD parameters are updated.

\subsection{Confidence Extrapolation (CE)}
\label{app:ce_impl}
CE is a training-free per-token forecasting module that models the evolution of a scalar confidence margin.
Let $\delta_i^{(t)} = \ell_{i,1}^{(t)}-\ell_{i,2}^{(t)}$ be the margin between top-1 and top-2 confidences at position $i$.
We use a 2D constant-velocity state model
$
x_i^{(t)} = \begin{bmatrix}\delta_i^{(t)} \\ \dot{\delta}_i^{(t)}\end{bmatrix},
\quad
x_i^{(t-1)} = A x_i^{(t)} + \epsilon,\quad
A=\begin{bmatrix}1&1\\0&1\end{bmatrix},
$
with Gaussian process noise $\epsilon\sim\mathcal{N}(0,Q)$ and observation $o_i^{(t)}=\delta_i^{(t)}+\eta$ with $\eta\sim\mathcal{N}(0,R)$.
We run a Kalman filter to estimate $(\hat{x}_i^{(t)},P_i^{(t)})$ and perform $h$-step prediction to obtain a conservative margin forecast
$
\hat{\delta}_{i,h}^{(t)} - z\sqrt{\mathrm{Var}(\delta_{i,h}^{(t)})},
$
where $z$ controls risk sensitivity.
This value is mapped to a conservative confidence proxy $\tilde{c}_i^{(t)}$ via a monotone transform (e.g., sigmoid calibration) and used as an additional trace signal and/or to choose a safe look-ahead horizon $h$.
CE adds only scalar operations per token per step and is negligible compared to the base model forward pass.

\subsection{Inference Integration}
\label{app:infer_integration}
At inference step $t$, we run one forward pass of $\mathcal{M}$, compute trace features for active tokens, optionally run CE to obtain forecast-based signals, and apply TSPD to decide which tokens to fix.
Fixed tokens are excluded from remasking and remain constant in subsequent steps.
We then apply the standard remasking schedule to the remaining active tokens and proceed to step $t-1$.
We terminate when all response tokens are fixed or the step budget is exhausted.

\begin{figure}[t]
    \centering
    \includegraphics[width=0.28\linewidth]{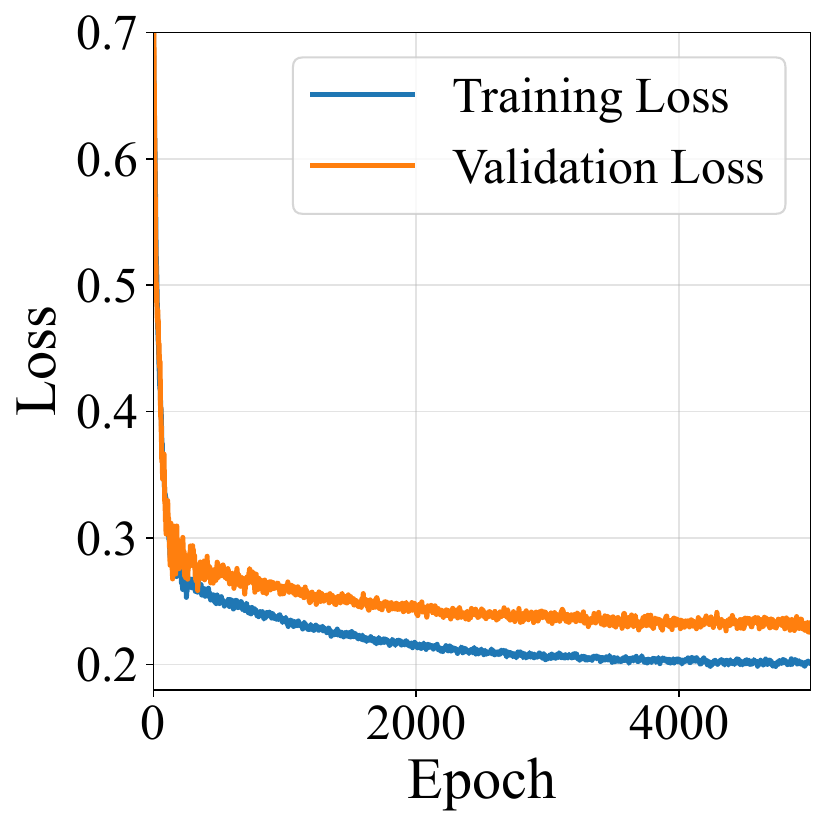}
    \caption{Learning curve of the TSPD. The learning curves illustrate the progression of training and validation loss across 5,000 epochs.}

    \vspace{-4mm}
    \label{curve}
\end{figure}

\subsection{Baselines and KV Caching}
\label{app:baselines_kv}
For Fast-dLLM and Credit Decoding, we enable their KV caching implementations and report their default settings.
For Learn2PD and Prophet, we follow their published inference procedures without KV caching.
When evaluating compatibility, we apply the same KV caching strategy to our method by caching the dLLM intermediate states used by the underlying forward passes, while keeping the controller and CE unchanged.
This isolates the gains from TSPD and CE and verifies orthogonality to system-level reuse.



\section{Training Curve of TSPD}
As shown in Figure~\ref{curve}, both the training loss (blue) and validation loss (orange) drop rapidly in the early epochs, decreasing from roughly $0.7$ to around $0.28$ within the initial phase.
After this sharp descent, the curves enter a slow refinement regime and decrease gradually over the remainder of the 5{,}000-epoch run, forming a near-flat tail.
Throughout training, the validation loss stays slightly above the training loss, indicating a small and stable generalization gap rather than severe overfitting.
The mild jitter in both curves reflects stochastic optimization noise, while the long, slowly decreasing tail suggests diminishing returns and that the model has effectively converged by the end of training.





\end{document}